\documentclass[bioengineering,article,accept,pdftex,moreauthors]{Definitions/mdpi} 

\usepackage{xcolor}

\firstpage{1} 
\pubvolume{1}
\issuenum{1}
\articlenumber{0}
\pubyear{2026}
\copyrightyear{2026}
\externaleditor{Muneer Ahmad} 
\datereceived{23 December 2025} 
\daterevised{11 February 2026} 
\dateaccepted{12 February 2026} 
\datepublished{ } 
\hreflink{https://doi.org/} 

\Title{Synthetic Melanoma Image Generation and Evaluation Using Generative Adversarial Networks}

\Author{Pei-Yu Lin 
 $^{1}$, Yidan Shen $^{2}$\orcidA{}, Neville Mathew $^{1}$\orcidB{}, Renjie Hu $^{3}$\orcidC{}, Siyu Huang $^{4}$, Courtney M. Queen $^{5}$\orcidG{}, \linebreak Cameron E. West
 $^{6}$\orcidE{}, Ana Ciurea $^{7}$\orcidF{} and George Zouridakis$^{1,}$$^{2}$$^{,8}$*\orcidD{}}

\AuthorNames{Pei-Yu Lin, Yidan Shen, Neville Mathew, Renjie Hu, Siyu Huang, Courtney M. Queen, Cameron E. West, Ana Ciurea, and George Zouridakis}

\address{%
$^{1}$ \quad Department of Engineering Technology, University of Houston, Sugar Land, TX 77479, 
 USA;  \linebreak plin22@cougarnet.uh.edu (P.-Y.L.);  namathew3@uh.edu (N.M.) 
 \\
$^{2}$ \quad Department of Electrical and Computer Engineering, University of Houston, Houston, TX 77204, USA; yshen20@uh.edu\\
$^{3}$ \quad Department of Information Science Technology, University of Houston, Sugar Land, TX 77479, USA; rhu7@uh.edu\\
$^{4}$ \quad School of Computing, Clemson University, Clemson, SC 29634, USA; siyuh@clemson.edu\\
$^{5}$ \quad Department of Public Health, Texas Tech University, Abilene, TX 79601, USA; courtney.m.queen@ttuhsc.edu\\
$^{6}$ \quad Department of Dermatology and Pathology, Texas Tech University, Abilene, TX 79601, USA; cameron.west@ttuhsc.edu\\
$^{7}$ \quad Department of Dermatology, University of Texas MD Anderson Cancer Center, Houston, TX 77030, USA; amciurea@mdanderson.org\\
$^{8}$ \quad Department of Biomedical Engineering, University of Houston, Houston, TX 77204, USA
}

\corres{Correspondence: zouridakis@uh.edu}

\abstract{
Melanoma is the most lethal form of skin cancer, and early detection is critical for improving patient outcomes. Although dermoscopy combined with deep learning has advanced automated skin-lesion analysis, progress is hindered by limited access to large, well-annotated datasets and by severe class imbalance, where melanoma images are substantially underrepresented. To address these challenges, we present the first systematic benchmarking study comparing four GAN architectures---DCGAN, StyleGAN2, and two StyleGAN3 variants (T and R)---for high-resolution ($512\times512$) melanoma-specific synthesis. We train and optimize all models on two expert-annotated benchmarks (ISIC 2018 and ISIC 2020) under unified preprocessing and hyperparameter exploration, with particular attention to R1 regularization tuning. Image quality is assessed through a multi-faceted protocol combining distribution-level metrics (FID), sample-level representativeness (FMD), qualitative dermoscopic inspection, downstream classification with a frozen EfficientNet-based melanoma detector, and independent evaluation by two board-certified dermatologists. StyleGAN2 achieves the best balance of quantitative performance and perceptual quality, attaining FID scores of 24.8 (ISIC 2018) and 7.96 (ISIC 2020) at $\gamma=0.8$. The frozen classifier recognizes 83\% of StyleGAN2-generated images as melanoma, while dermatologists distinguish synthetic from real images at only 66.5\% accuracy (chance = 50\%), with low inter-rater agreement ($\kappa = 0.17$). In a controlled augmentation experiment, adding synthetic melanoma images to address class imbalance improved melanoma detection AUC from 0.925 to 0.945 on a held-out real-image test set. These findings demonstrate that StyleGAN2-generated melanoma images preserve diagnostically relevant features and can provide a measurable benefit for mitigating class imbalance in melanoma-focused machine learning pipelines.}


\begin{document}

\section{Introduction}

Melanoma accounts for only a small fraction of skin cancer diagnoses (about 6\%) \cite{ACS2023CancerFacts}, yet it is responsible for the majority of skin-cancer-related deaths \cite{ACS2023CancerFacts, Siegel2022CancerStats}. In the United States, an estimated 186,680 melanoma cases were diagnosed in 2023, resulting in 7990 deaths \cite{ACS2023CancerFacts}. Because the 5-year survival rate exceeds 99\% when melanoma is detected early \cite{ACS2023CancerFacts}, timely and accurate recognition of suspicious lesions is clinically critical. However, melanoma remains difficult to detect reliably due to substantial intra-class variability (e.g., color, texture, borders) and frequent atypical presentations that overlap visually with benign lesions \cite{Luke2017MelanomaTherapy}. Automated screening systems based on dermoscopic criteria such as the 7-point checklist have shown promise for early detection \cite{wadhawan2011implementation, situ2010modeling, zouridakis2020methods}, but their performance depends critically on the availability of large, diverse training datasets.

Recent advances in machine learning have improved automated dermoscopic image analysis, but progress is constrained by limited access to high-quality expert annotations and by severe class imbalance \cite{Zareen2025Innovations, yaopaper, raza2025neuromoe}: melanoma images are typically far fewer than non-melanoma cases in commonly used datasets. This imbalance can bias learned decision boundaries and reduce generalization, especially when models are deployed across acquisition settings, devices, and patient populations.

Generative modeling has therefore emerged as a promising direction to mitigate data scarcity and imbalance. In particular, Generative Adversarial Networks (GANs) can synthesize dermoscopic images that resemble real lesions and may enrich melanoma-specific variability for training and benchmarking \cite{Innani2023SkinLesionSegGAN, LaSalvia2022SensorsSkinGAN}. Nevertheless, many prior approaches rely on relatively early or constrained generator designs that either produce limited-resolution images failing to preserve fine-grained dermoscopic cues, or depend heavily on feature-space metrics alone, which may not reflect clinical usefulness or downstream recognizability~\cite{Bioengineering2025DCGANMelanoma, FumagalGonzalez2025PGGANMelanoma}. Moreover, melanoma-focused synthesis and systematic cross-architecture comparisons remain relatively underexplored.

In this study, we present the first systematic benchmark comparing four GAN-based architectures for high-resolution ($512\times512$) melanoma image synthesis, while these architectures are well-established in the general computer vision literature, their relative performance for melanoma-specific synthesis---where preservation of fine-grained dermoscopic features is essential---has not been comprehensively evaluated. We train and optimize all models on two expert-annotated datasets (ISIC 2018 and ISIC 2020) under unified preprocessing and hyperparameter exploration, enabling direct cross-architecture comparison. Beyond standard generative metrics, we employ a multi-faceted evaluation protocol combining distribution-level assessment (FID), sample-level representativeness (FMD), qualitative dermoscopic inspection, downstream validation using a frozen EfficientNet-based melanoma classifier, and independent assessment by board-certified dermatologists. This combination of systematic comparison and clinically grounded evaluation addresses a gap in the literature, where prior studies often evaluate single architectures or rely solely on feature-space metrics.

The main contributions of this work are threefold: (1) a systematic cross-architecture comparison of DCGAN, StyleGAN2, and StyleGAN3 variants (T and R) for melanoma-specific image synthesis under consistent experimental conditions, with empirical evidence that StyleGAN2 provides the best balance of quantitative performance, perceptual quality, and artifact avoidance; (2) domain-specific hyperparameter optimization, particularly regarding R1 regularization strength ($\gamma$), with practical guidance for melanoma synthesis; and (3) a multi-faceted evaluation protocol combining distribution-level metrics, sample-level representativeness, downstream classifier validation, and independent assessment by two board-certified dermatologists. This demonstrated that 
 83\% of StyleGAN2-generated images are recognized as melanoma by a strong external classifier and that expert dermatologists distinguish synthetic from real images at only 66.5\% accuracy.

\section{Related Work}

\subsection{GANs in Medical Imaging} 
The prevalence of severe class imbalance has motivated a shift from traditional geometric data augmentation toward generative modeling techniques. Early studies employed DCGANs to generate dermoscopic images for skin lesion classification but achieved limited realism due to low image resolution and insufficient preservation of fine-grained diagnostic structures \cite{Bissoto2019SkinSynthesisGAN}. 

More recently, Behara et al.\ \cite{Behara2023DCGAN} proposed an improved DCGAN classifier for skin lesion synthesis, demonstrating that careful hyperparameter tuning and image preprocessing can enhance DCGAN performance on dermatological datasets. Conditional GANs have also been applied to melanoma-specific tasks. For example, Ali et al.\ \cite{Ali2023IoMTcGAN} utilized cGANs for melanoma lesion segmentation in IoMT-based systems, illustrating the versatility of adversarial frameworks across different medical imaging objectives.

Similar limitations were observed in other GAN-based dermatology studies, where low-resolution synthesis constrained clinical applicability \cite{FridAdar2018LiverGAN}. To address resolution and stability issues, Progressive Growing GANs (PGGANs) were introduced into medical imaging. PGGANs enabled stable synthesis of high-resolution dermoscopic images, significantly improving visual fidelity and structural consistency compared to DCGANs \cite{Karras2018PGGAN, Bissoto2019BiasSkin}. Subsequently, StyleGAN-based models enabled fine-grained manipulation of lesion morphology and appearance \cite{Karras2019StyleGAN}. StyleGAN-ADA demonstrated strong performance on limited medical datasets by dynamically adapting data augmentation strategies \cite{Karras2020LimitedDataGAN}. Beyond dermatology, FundusGAN has been applied to retinal imaging, preserving complex vascular structures and enabling effective augmentation for ophthalmic disease classification \cite{Costa2018FundusInterpretability}. 

\subsection{Emerging Alternatives: Diffusion Models}
Denoising Diffusion Probabilistic Models (DDPMs) have recently emerged as a powerful alternative to GANs for image synthesis. Dhariwal and Nichol \cite{Dhariwal2021Diffusion} demonstrated that diffusion models can surpass GANs on standard benchmarks such as ImageNet, achieving state-of-the-art FID scores through architectural improvements and classifier guidance. Latent diffusion models further improved computational efficiency by operating in compressed latent spaces \cite{Rombach2022LatentDiffusion}, enabling high-resolution synthesis with reduced memory requirements. Akrout et al.\ \cite{Akrout2024DiffusionSkin} evaluated diffusion-based augmentation for skin disease classification, finding that synthetic images can match classifier performance when appropriately curated. Farooq et al.\ \cite{Farooq2024DermT2IM} proposed Derm-T2IM, a text-to-image framework using Stable Diffusion to generate melanoma and benign lesion images from natural language prompts. More recently, Wang et al.\ \cite{Wang2024MinorityDiffusion} applied diffusion-based augmentation specifically to address underrepresentation of minority subgroups in skin lesion datasets. 

Despite these promising developments, diffusion models for dermatology remain in early stages relative to GAN-based approaches, with limited systematic evaluation on melanoma-specific synthesis tasks. Furthermore, diffusion models introduce different tradeoffs: while they offer improved training stability, they typically require substantially longer inference times, and their ability to preserve fine-grained dermoscopic features has not been extensively validated. The present study therefore focuses on GAN architectures, which remain the most thoroughly characterized family for medical imaging synthesis, while acknowledging diffusion-based methods as a promising direction for future investigation.

\subsection{Synthetic Data Generation for Melanoma Imaging}

Severe data scarcity and class imbalance in melanoma imaging have motivated generative modeling as an alternative to purely geometric augmentation. However, early applications of classic GAN backbones often struggled to faithfully reproduce fine-grained diagnostic structures, especially at higher resolutions \cite{Bioengineering2025DCGANMelanoma}. To improve resolution and fidelity, progressive-growing strategies have been explored. Fumagal-Gonzalez et al.\ \cite{FumagalGonzalez2025PGGANMelanoma} employed PGGAN for melanoma synthesis and investigated how different real-to-synthetic ratios affect downstream melanoma detection, while PGGAN yielded visually richer samples, the authors observed that performance gains were not monotonic with increasing synthetic data and noted occasional suboptimal generations under practical training and hardware constraints, underscoring the remaining stability and consistency challenges of high-resolution GAN training. More recently, Abbasi et al.\ \cite{Abbasi2025DiffusionMelanoma} fine-tuned a pre-trained Stable Diffusion model with LoRA and reported melanoma image generation with improved fine details, suggesting that large diffusion backbones can better capture complex lesion appearance. At the same time, diffusion models raise new practical considerations such as computational footprint and the need for careful domain validation when deployed for medical data augmentation. Alongside model development, Luschi et al.\ \cite{Luschi2025ValidationProtocolMelanoma} proposed a holistic validation protocol for GAN-generated melanoma images that integrates objective computational metrics with structured expert assessment. Importantly, this line of work primarily advances how to validate synthetic melanoma images; it does not provide a controlled, large-scale benchmark of modern GAN architectures under consistent training~protocols.

In summary, although generative models have advanced from DCGAN to PGGAN and, more recently, diffusion-based approaches, the literature still lacks systematic and controlled comparisons of state-of-the-art GANs for melanoma image synthesis. The present study addresses this gap.

\section{Methods}

\subsection{Generative Models}
Generative Adversarial Networks (GANs) \cite{Goodfellow2014GAN} consist of two neural networks, a generator ($G$) and a discriminator ($D$), trained simultaneously in a competitive manner. The generator aims to produce synthetic data whose distribution closely resembles that of the real data, while the discriminator attempts to distinguish between real and generated samples. The training process can be formulated as a minimax game, where both $G$ and $D$ iteratively optimize their respective objectives. GAN training can be expressed mathematically as the following optimization problem \cite{Bastien2012Theano}:
\[
\min_{G} \max_{D} V(D,G)
= \mathbb{E}_{x \sim p_{\text{data}}(x)} \left[ \log D(x) \right]
+ \mathbb{E}_{z \sim p_{z}(z)} \left[ \log \left( 1 - D(G(z)) \right) \right]
\]

In this formulation, the generator $G$ aims to minimize the objective function by creating data that are indistinguishable from real samples, while the discriminator $D$ seeks to maximize it by accurately distinguishing real data $x$ from generated samples $G(z)$. 

The term $\mathbb{E}_{x} \left[ \log D(x) \right]$ quantifies the discriminator's success in recognizing real data, whereas $\mathbb{E}_{z} \left[ \log \left( 1 - D(G(z)) \right) \right]$ measures its ability to identify generated data as fake, where $z$ denotes the input noise vector.

\begin{figure}[H]
\includegraphics[width=9.0 cm]{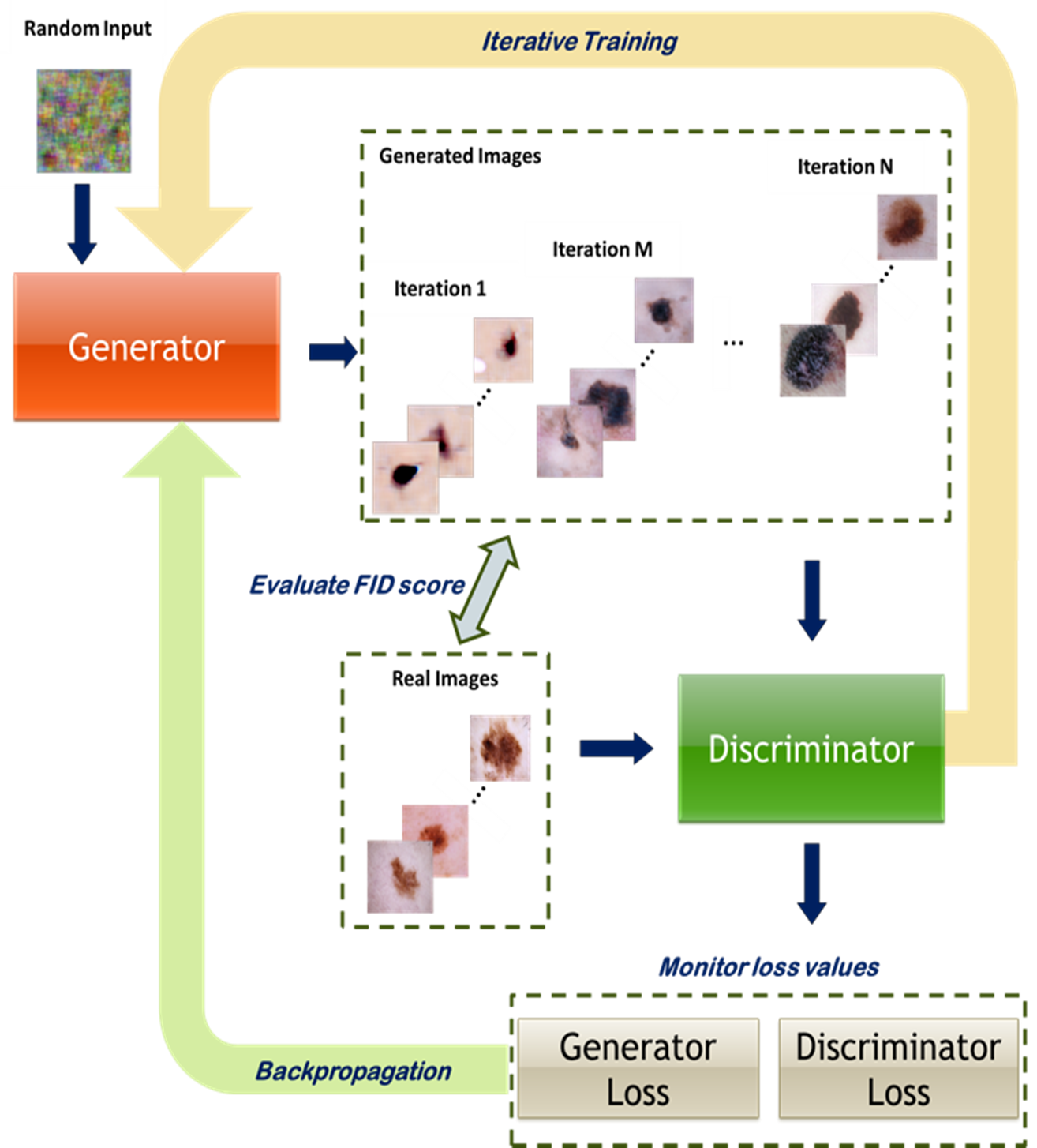}
\caption{Iterative GAN training: the generator and discriminator undergo concurrent adversarial training. \label{fig1}}
\end{figure}

This adversarial process encourages the generator to improve the quality of synthesized data and the discriminator to enhance its classification accuracy, thereby improving overall GAN performance. The two networks are trained simultaneously through this iterative adversarial procedure, as illustrated in \cref{fig1}.

\begin{figure}[H]
\includegraphics[width=10.0 cm]{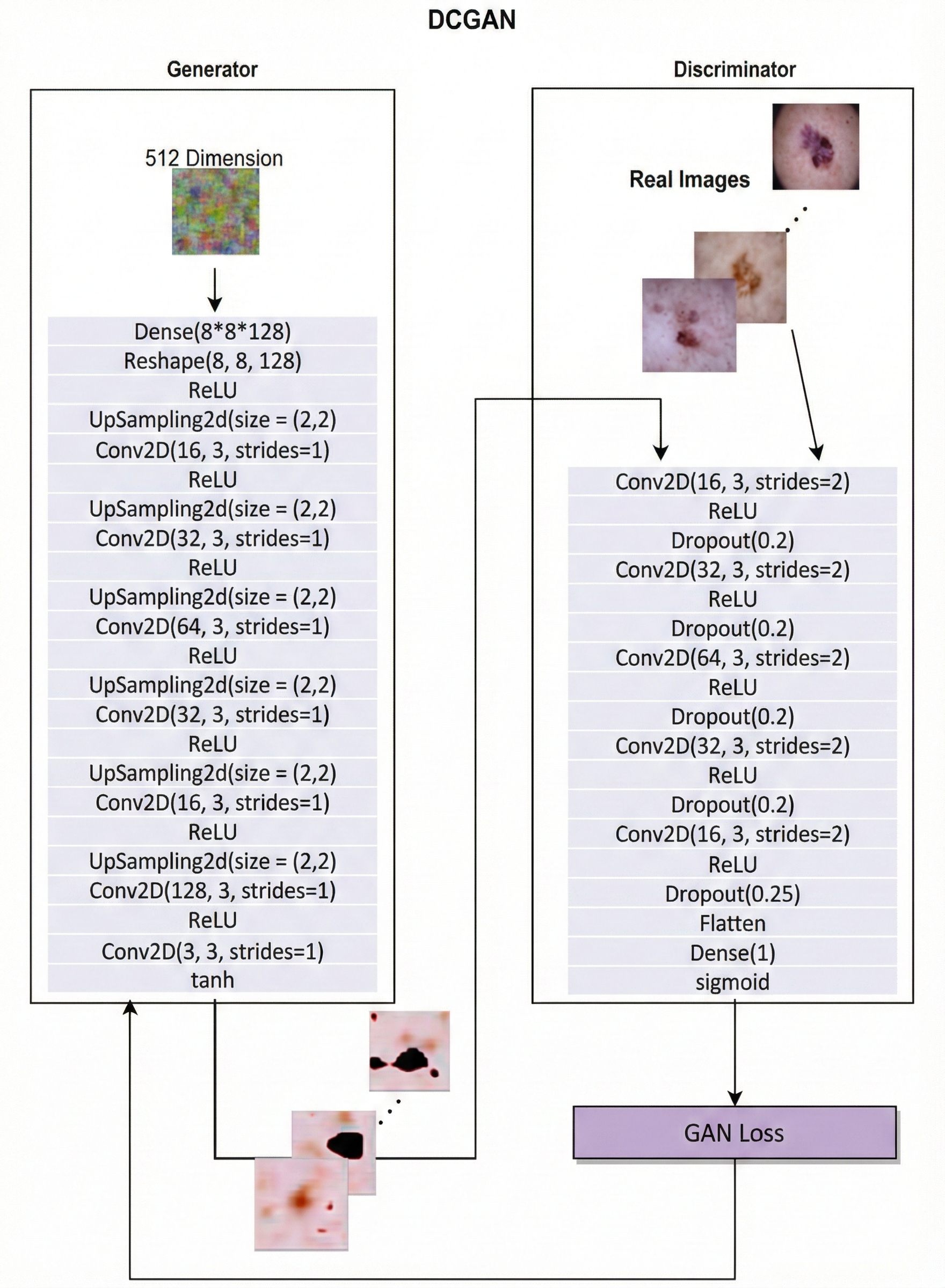}
\caption{
 DCGAN architecture showing the generator and discriminator networks, including layer configuration, upsampling and strided convolutions, and output activations. \label{fig2}}
\end{figure}

The DCGAN model:
Deep Convolutional Generative Adversarial Networks (DCGANs) employ convolutional neural networks in both the generator and discriminator \cite{Radford2015DCGAN} to capture spatial image structure, including edges, textures, and object relationships. Our implementation is shown in \cref{fig2} along with the specific parameters used.

\begin{figure}[H]
\includegraphics[width=13cm]{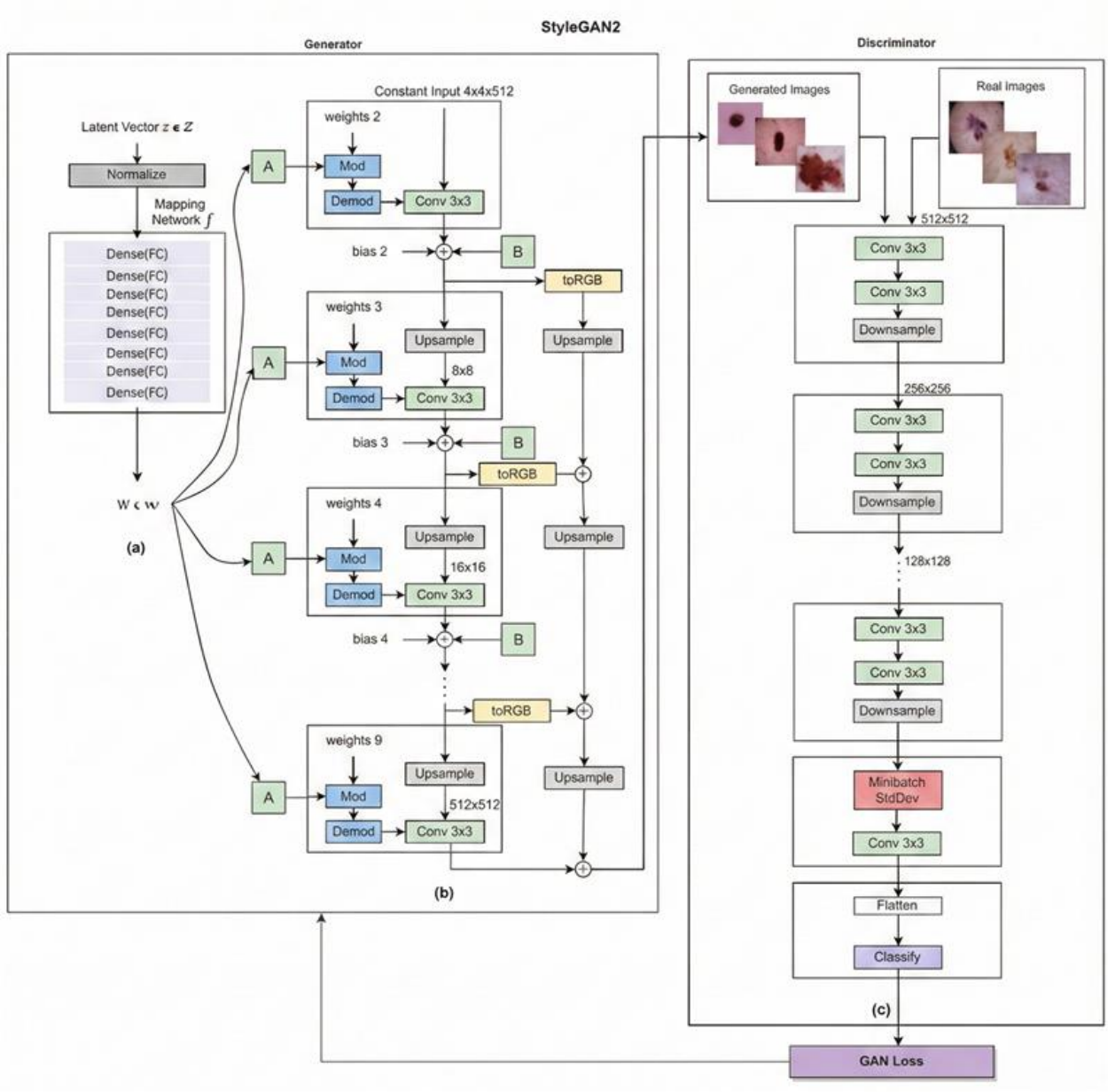}
\caption{
StyleGAN2 architecture illustrating the generator and discriminator networks. In the generator, \textbf{(a)} denotes the learned affine transformation used for style modulation, \textbf{(b)} represents the learned per-channel scaling factor, and $\oplus$ indicates element-wise addition. In the discriminator, \textbf{(c)} denotes the final classification stage. \label{fig3}}
\end{figure}

The StyleGAN models: The Style-based Generative Adversarial Network (StyleGAN) disentangles different aspects of image generation, such as content and style, through Adaptive Instance Normalization (AdaIN) layers that enable precise control over the output. The general architecture is shown in \cref{fig3}. A random noise input vector ($z$) passes through the mapping network and is transformed into an intermediate latent space. The synthesis network (\cref{fig3}b) progressively adds layers that increase the resolution of generated images (\cref{fig3}c). The discriminator takes an image (real or generated) as input and produces a single output value representing the probability that the input image is real. The original StyleGAN uses a combined approach for content and style manipulation that limits independent adjustment; accordingly, it was not included in our evaluation. Instead, we selected StyleGAN2, which introduces a modular architecture that separates content and style \cite{Karras2020StyleGAN2, Karras2019StyleGAN}. Both models utilize progressive growing for high-resolution image generation, but StyleGAN2 employs enhanced upsampling techniques for sharper details. StyleGAN3 \cite{Karras2021AliasFree} is the latest version in the series and addresses the issue of ``texture sticking,'' where repetitive patterns appear in some StyleGAN2-generated images. It relies on an alias-free generator architecture that uses Fourier features to represent image content.

\subsection{Experimental Setup}
\subsubsection{Compute Cluster}
Experiments were conducted on a DGX-2 system with 16 NVIDIA Tesla V100 GPUs (combined processing power of 2 petaflops), 512 GB of total GPU memory, 1.5 TB of NVMe storage, 15 TB of SATA storage, and the {NVIDIA CUDA software, v12.2} 
 stack. Four GPUs, each with 32 GB memory, were used for all model training.

\subsubsection{Datasets}
The ISIC 2018 dataset \cite{Codella2019ISIC2018Challenge}, provided by the International Skin Imaging Collaboration (ISIC), comprises 10,015 high-quality images representing seven different skin lesion types, including 1113 melanomas. The images are highly variable in terms of lighting, resolution, and lesion appearance. Images are annotated with a specific diagnosis and may include lesion localization, patient age, and sex. The second dataset, ISIC 2020 \cite{isic2020dataset}, contains 33,126 dermoscopic skin lesion images, including 7227 melanomas, all with associated metadata. 

\subsubsection{Data Preprocessing}
For model training, we used 1061 melanoma images from the ISIC 2018 dataset after excluding 52 images with excessive artifacts, poor focus, or non-standard framing. All images were resized from their original dimensions to $512 \times 512$ pixels. 

For DCGAN training, we employed bilateral filtering \cite{Tomasi1998Bilateral, Gavaskar2018FastBilateral} to reduce noise while preserving edge structures, followed by image normalization across the dataset. The training set was augmented to 8488 images through rotations (90\textdegree, 180\textdegree, and 270\textdegree) and horizontal flipping. For StyleGAN2 and StyleGAN3 training, we applied only horizontal flipping to a random subset of images, as these architectures incorporate internal augmentation mechanisms that reduce the need for extensive external augmentation. This difference in augmentation strategy reflects architecture-specific best practices rather than an experimental variable; however, it should be considered when interpreting cross-architecture comparisons. Augmentation improved model robustness and reduced overfitting to specific orientations of melanoma patterns. Model parameters were further optimized using 7227 images from the ISIC 2020 dataset with the same preprocessing procedure.

\subsubsection{Model Parameter Exploration}
For the DCGAN model, we configured the kernel size to $(3, 3)$ and used LeakyReLU activation in the convolution layers, with tanh and sigmoid activations in the output layers of the generator and discriminator, respectively. The hyperparameters were set as follows: noise vector size of 512, batch size of 32, and maximum training iterations of 300,000---training images were repeatedly sampled from the melanoma dataset until this threshold was reached. We used the TruncatedNormal initializer, binary cross-entropy loss function, and the Adam optimizer with a learning rate of 0.0002 and $\beta_1 = 0.5$ \cite{Mutepfe2021SkinLesionGAN}. We explored output resolutions ranging from $256 \times 256$ to $512 \times 512$ pixels, several dropout rates, different filter values for each layer, and the effects of batch normalization in both the generator and discriminator. 

For the StyleGAN2 and StyleGAN3 models, we fixed the batch size at 32 and focused on optimizing the R1 regularization weight ($\gamma$), which plays a critical role in stabilizing training. We explored $\gamma \in \{0.8, 1.6, 8.0, 10.0\}$ based on recommendations in \cite{Karras2021AliasFree}. StyleGAN3 was trained using two configurations: StyleGAN3-R (rotation and translation equivariance), designed to minimize positional bias and improve rotational consistency~\cite{Karras2021AliasFree}; and StyleGAN3-T (translation equivariance), which emphasizes realistic textures and fine details \cite{Karras2021AliasFree}. The maximum number of training images was set to 6,800,000, with images repeatedly sampled from the melanoma training dataset until this count was reached.

\subsection{Model Evaluation}
We used the Fr\'echet Inception Distance (FID) \cite{Heusel2017TTUR}, which quantifies similarity between the distributions of real and synthetic images by comparing deep features extracted from the pre-trained Inception V3 network \cite{Szegedy2016InceptionV3}. FID is computed as:
\[
\text{FID} = \|\mu_r - \mu_g\|^2 + \text{Tr}\left(\Sigma_r + \Sigma_g - 2(\Sigma_r \Sigma_g)^{1/2}\right)
\]
where $(\mu_r, \Sigma_r)$ and $(\mu_g, \Sigma_g)$ are the mean and covariance of the real and generated feature distributions, respectively. Lower FID indicates greater similarity to the real data distribution. However, FID scores are not directly interpretable in terms of human perception, and a lower score does not always guarantee that generated images will appear more realistic or prove useful in practical applications. 

In addition to FID, we monitored generator and discriminator loss values across parameter settings to ensure loss convergence and identify potential model collapse, training instability, or discriminator dominance.

To address limitations of single-metric evaluation, recent work has emphasized the importance of multi-faceted assessment for synthetic medical images. Abdusalomov~\mbox{et al.\ \cite{Abdusalomov2023SyntheticGAN}} highlighted that existing metrics primarily evaluate distributional similarity but may fail to capture whether synthetic images preserve medically relevant features or introduce artifacts affecting downstream utility. Following these recommendations, we complement FID with the Fr\'echet Medoid Distance (FMD), which measures the distance from each generated sample to the medoid (most central sample) of the real distribution in feature space, providing a sample-level measure of representativeness that is more sensitive to mode collapse than distribution-level metrics. We further include qualitative dermoscopic inspection to identify clinical feature preservation, and downstream classifier evaluation to test retention of discriminative cues.

\section{Results}
\subsection{FID and FMD Performance Analysis}

To comprehensively evaluate the quality of generated images, we jointly consider the Fr\'echet Inception Distance (FID) and the Fr\'echet Medoid Distance (FMD), which quantify complementary aspects of generative performance. FID measures how closely the global feature distribution of generated samples matches that of real images by computing the Fr\'echet distance between Gaussian approximations in Inception feature space; lower values indicate better overall realism and diversity. In contrast, FMD evaluates sample-level representativeness by measuring distances to medoid real samples in feature space, making it more sensitive to mode collapse and local mismatches; again, lower values indicate better~performance.

Table \ref{tab:performance_models} reports the results for all models under $\gamma=8$, which serves as a common reference point for cross-architecture comparison. In terms of FID, StyleGAN3-R achieves the lowest score (26.47), with StyleGAN2 close behind (31.58), both substantially outperforming DCGAN (66.49), while StyleGAN3-T performs markedly worse (246.42). 

Interestingly, StyleGAN3-T presents a divergent pattern: while achieving the worst FID score (246.42), it obtains the lowest FMD (41.37). This apparent discrepancy reflects the complementary nature of these metrics. FID evaluates distributional similarity by comparing Gaussian approximations of feature distributions, penalizing models that fail to capture the full diversity of the real data. FMD, in contrast, measures sample-level representativeness by computing distances to medoid real samples, rewarding individual images that closely resemble typical real examples regardless of overall diversity. The combination of poor FID and strong FMD for StyleGAN3-T suggests limited mode coverage: the model generates samples that individually resemble real melanomas but fails to capture the full morphological diversity present in the training distribution---a pattern consistent with partial mode collapse. StyleGAN2 and StyleGAN3-R achieve strong performance on both metrics (FID: 31.58 and 26.47; FMD: 50.08 and 49.21, respectively), indicating both distributional fidelity and sample-level representativeness. DCGAN performs poorly on both metrics, reflecting limited capacity for high-resolution medical image synthesis.

\begin{figure}[H]
\includegraphics[width=13 cm]{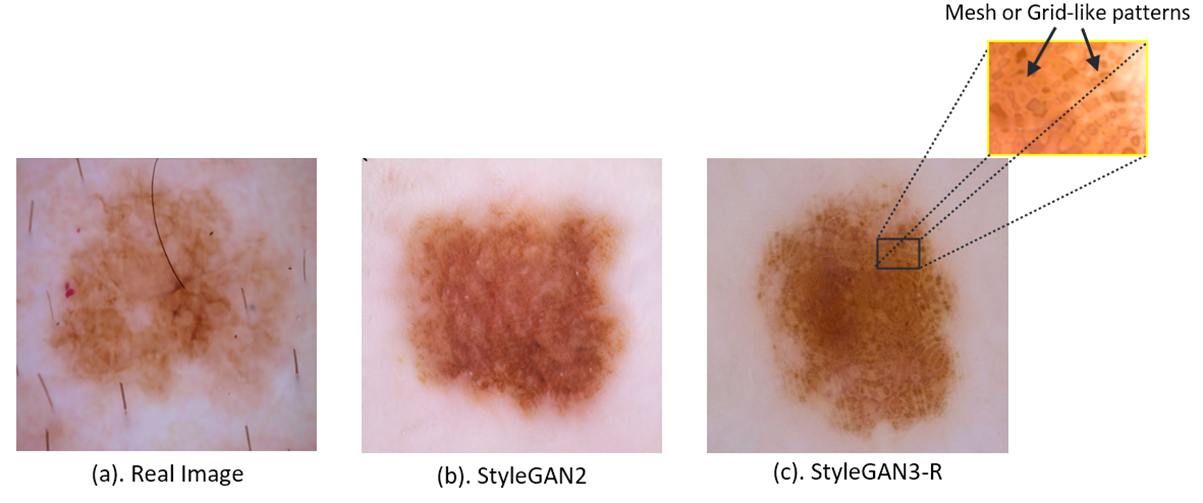}
\caption{(\textbf{a}) Real melanoma and images produced by (\textbf{b}) StyleGAN2 and (\textbf{c}) StyleGAN3-R. The zoomed inset highlights mesh- or grid-like artifacts present in StyleGAN3-R outputs. \label{fig6}}
\end{figure}

Although StyleGAN3-R achieves the lowest FID at $\gamma = 8$, qualitative inspection of generated samples (\cref{fig6}) reveals prominent mesh- or grid-like artifacts. These patterns violate fundamental realism requirements for dermoscopic images and render the outputs unsuitable for medical applications, yet they are not adequately penalized by either feature-space metric. Visual inspection of 100 randomly selected StyleGAN3-R samples revealed such artifacts in approximately 60\% of images. For this reason, we select StyleGAN2 as the most reliable model overall, balancing strong quantitative performance with stable perceptual quality.

\begin{figure}[H]
\includegraphics[width=13.5 cm]{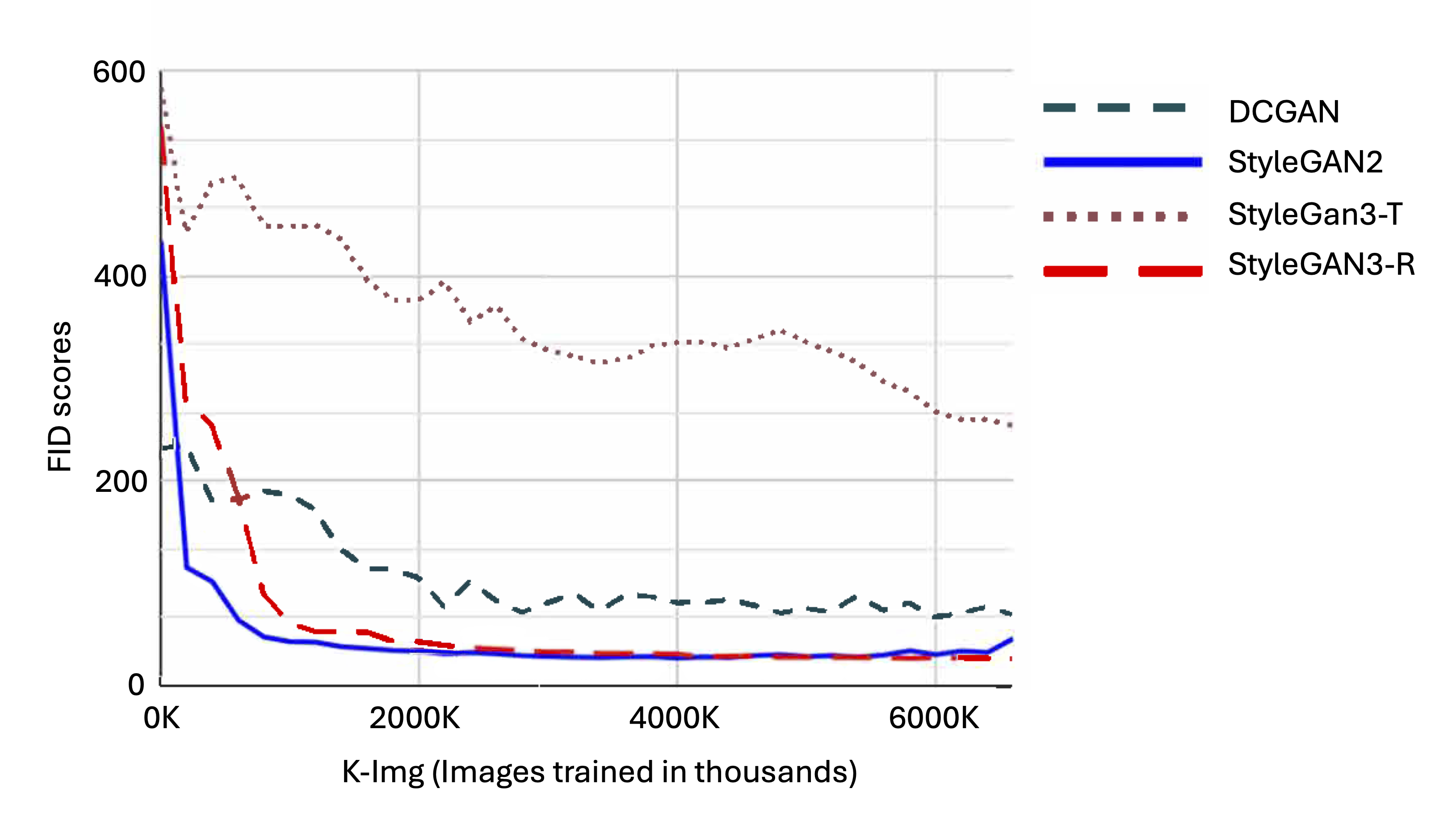}
\caption{FID score 
 as a function of training images (thousands) for each architecture. StyleGAN2 shows consistent improvement with additional data, while StyleGAN3-T plateaus at substantially higher FID values. \label{fig4}}
\end{figure}

Having identified StyleGAN2 as the preferred architecture, we investigated the effect of R1 regularization strength ($\gamma$) on its performance. Figure \ref{fig4} shows the evolution of FID with respect to training set size, demonstrating that StyleGAN2 benefits consistently from additional data, indicating robust scaling behavior. Furthermore, \cref{tab:fid_stylegan2_gamma} demonstrates that smaller $\gamma$ values yield better FID scores on both ISIC 2018 and ISIC 2020, with $\gamma=0.8$ producing the best results (24.8 and 7.96, respectively). This finding suggests that lighter regularization is preferable for melanoma synthesis, likely because the relatively homogeneous dermoscopic domain requires less aggressive smoothing of the discriminator's~gradients.

\begin{table}[H]
\caption{Performance comparison across architectures at $\gamma=8$. Lower FID and FMD values correspond to better performance ($\downarrow$). \label{tab:performance_models}}
\centering
\begin{tabularx}{\textwidth}{CCCCC}
\toprule
\textbf{Metric} & \textbf{DCGAN} & \textbf{StyleGAN2} & \textbf{StyleGAN3-T} & \textbf{StyleGAN3-R} \\
\midrule
FID $\downarrow$ & 66.49 & 31.58 & 246.42 & 26.47 \\
FMD $\downarrow$ & 695.93 & 50.08 & 41.37 & 49.21 \\
\bottomrule
\end{tabularx}
\end{table}
\unskip

\begin{table}[H]
\caption{FID scores for StyleGAN2 under different R1 regularization strengths ($\gamma$)
on the ISIC 2018 and ISIC 2020 datasets. Lower $\gamma$ values yield lower FID scores ($\downarrow$). \label{tab:fid_stylegan2_gamma}}
\centering
\begin{tabularx}{\textwidth}{CCC}
\toprule
\textbf{$\boldsymbol{\gamma}$} & \textbf{FID (ISIC 2018)} & \textbf{FID (ISIC 2020)} \\
\midrule
0.8  & 24.8 & 7.96 \\
1.6  & 27.4 & 9.48 \\
8.0  & 31.6 & 9.91 \\
10.0 & 33.2 & 10.4 \\
\bottomrule
\end{tabularx}
\end{table}
\unskip

\unskip

\subsection{Image Generation}
\begin{figure}[H]
\includegraphics[width=13.5 cm]{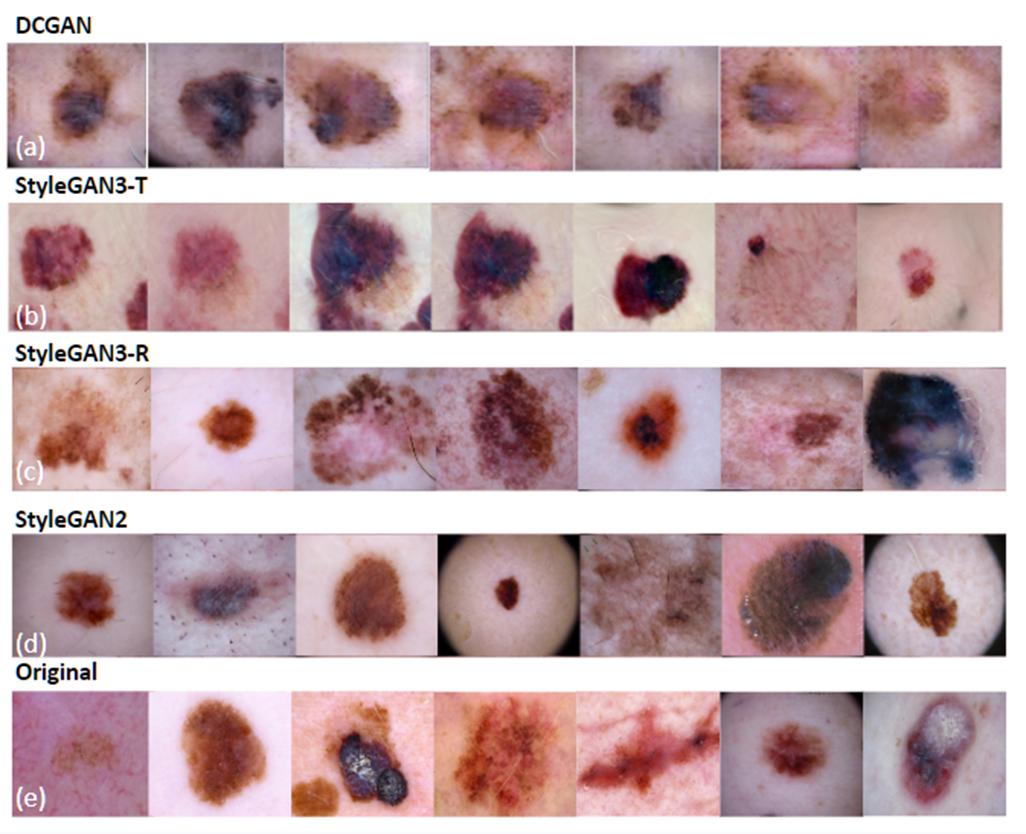}
\caption{Representative synthetic images from (\textbf{a}) DCGAN, (\textbf{b}) StyleGAN3-T, (\textbf{c}) StyleGAN3-R, and (\textbf{d}) StyleGAN2, compared with (\textbf{e}) real melanoma images used for training. \label{fig5}}
\end{figure}

Each model produced 1000 synthetic melanoma images per parameter configuration. Figure~\ref{fig5} shows representative examples from each model alongside real melanomas used for training. We evaluated the quality of synthetic images by examining the presence of characteristic dermoscopic features captured by the 7-point checklist \cite{Argenziano1998ABCD7Point}, which dermatologists use for melanoma diagnosis.

Despite training on over 6 million image presentations, DCGAN and StyleGAN3-T outputs lack the fine details expected in melanoma images, consistent with their elevated FID scores. In contrast, images produced by StyleGAN2 and StyleGAN3-R consistently exhibit the high-quality dermoscopic details present in real melanomas, including pigment network patterns, color variegation, and border irregularity. However, as noted above, StyleGAN3-R generates a substantial proportion of images with mesh- or grid-like artifacts (\cref{fig6}), attributable to the model's strict enforcement of translation and rotation equivariance constraints.

Overall, among all four architectures evaluated, StyleGAN2 produced the most realistic melanoma images in terms of the major and minor dermoscopic features of the 7-point checklist---an assessment corroborated by its strong FID scores.

\subsection{Computational Cost and Parameter Size}
We evaluated the computational efficiency of different GAN architectures in terms of training time and model size (\cref{tab:model_cost}). Using the ISIC 2018 dataset, DCGAN required 0.9 h to train, StyleGAN2 required 2.8 h to reach its optimal FID score, and both StyleGAN3-R and StyleGAN3-T required 9.2 h each.

The parameter count is reported as $G + D$, where $G$ and $D$ denote the number of parameters (in millions) in the generator and discriminator, respectively. DCGAN has a substantially smaller model size (5M total), while StyleGAN-based models employ larger networks (54--59M total), resulting in higher computational cost. Notably, StyleGAN2 achieves the best quality--efficiency tradeoff: it requires only 30\% of the training time of StyleGAN3 variants while producing superior or comparable output quality.

\begin{table}[H]
\caption{Model size and training time for each GAN architecture on ISIC 2018. \label{tab:model_cost}}

\begingroup
\small  

\centering
\begin{tabularx}{\textwidth}{lCCCC}
\toprule
 & \textbf{DCGAN} & \textbf{StyleGAN2} & \textbf{StyleGAN3-T} & \textbf{StyleGAN3-R} \\
\midrule
Parameters (M), $G + D$ & 4 + 1  & 30 + 29  & 25 + 29 & 25 + 29 \\
Training time (hours) &  0.9 & 2.8 & 9.2 & 9.2 \\
\bottomrule
\end{tabularx}

\endgroup
\end{table}

\section{Downstream Evaluation}

\subsection{Evaluation Model: External Skin Lesion Classifier}

To test whether synthetic melanoma images retain discriminative characteristics beyond feature-space similarity scores, we employed a strong external skin-lesion classifier as a downstream evaluator \cite{ha2020identifyingmelanomaimagesusing}. This model is an EfficientNet-B6-based ensemble developed for the SIIM-ISIC Melanoma Classification Challenge, where it achieved an AUC of 0.9490 on the private leaderboard, ranking among the top solutions. This strong baseline performance on real dermoscopic data makes it a rigorous test of whether synthetic images preserve melanoma-discriminative features. For consistency with our experimental setting, we map the model's outputs into two classes: melanoma and benign. 

\subsection{Downstream Evaluation I: Recognizability Under a Frozen Classifier}

\begin{figure}[H]
\subfloat[\centering]{\includegraphics[width=6.8cm]{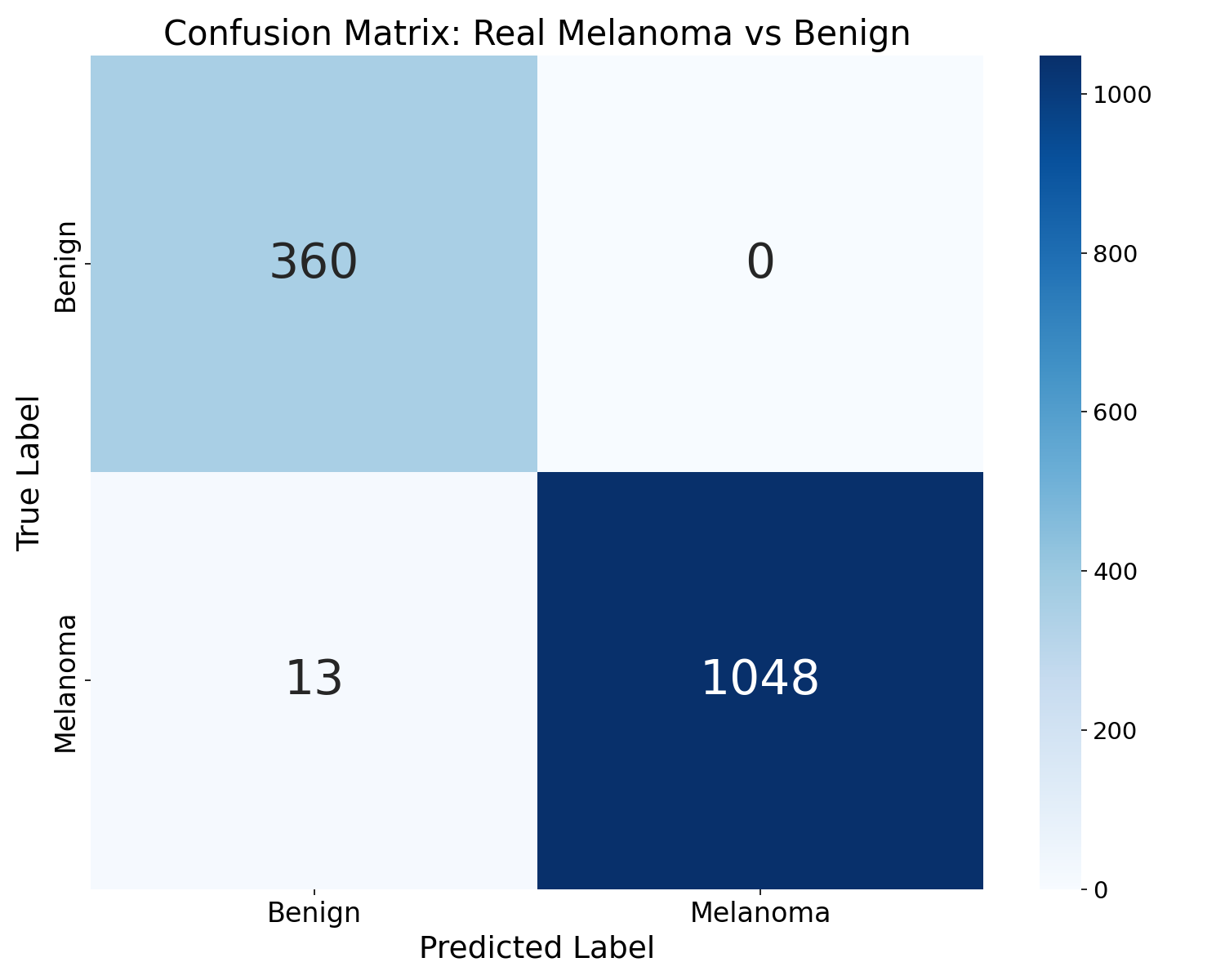}}
\subfloat[\centering]{\includegraphics[width=6.8cm]{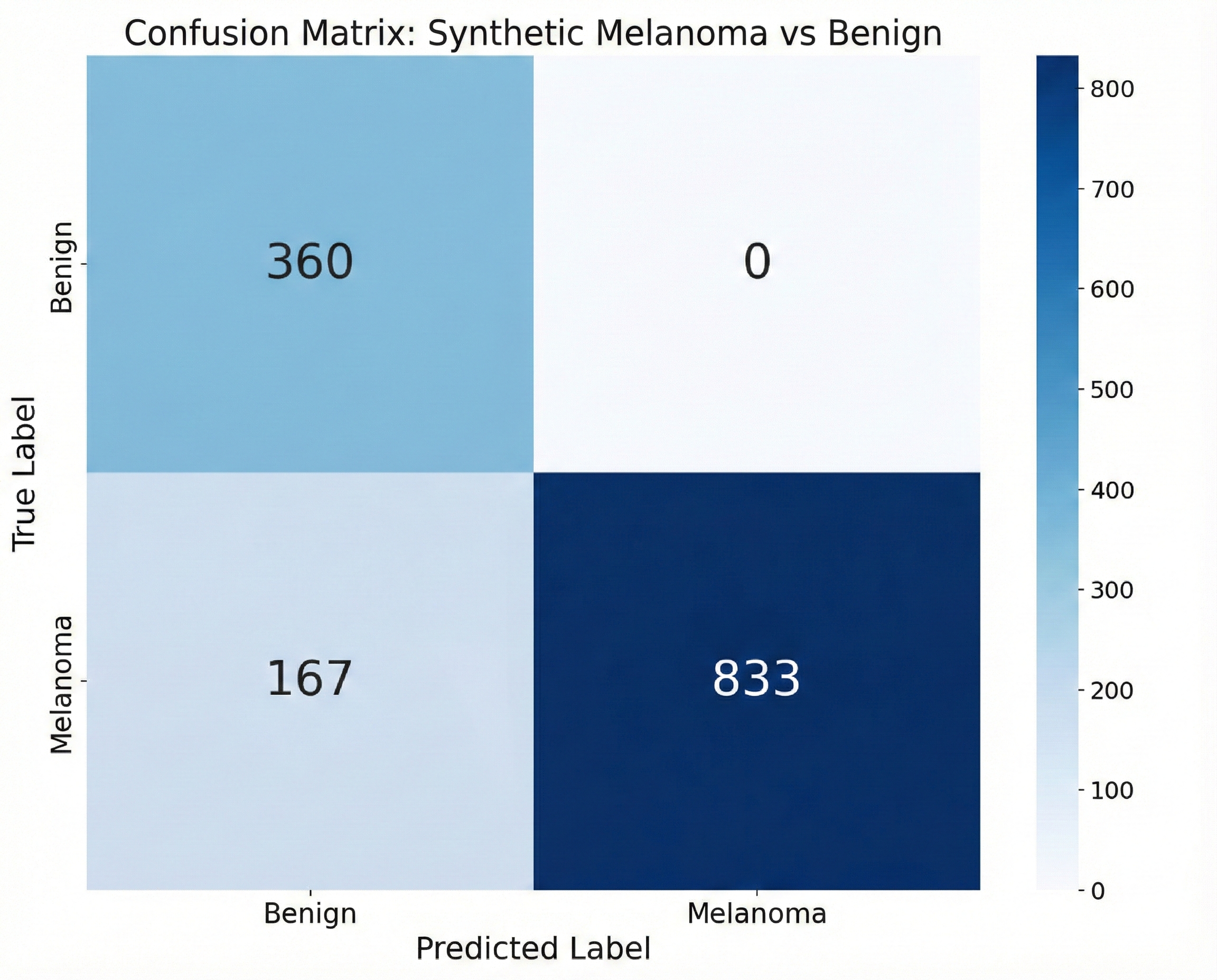}}
\caption{Confusion matrices for the frozen external classifier on (\textbf{a}) the Real set (real melanoma vs.\ real benign) and (\textbf{b}) the Synthetic set (StyleGAN2-generated melanoma vs.\ real benign). The classifier recognizes 83\% of synthetic melanomas as melanoma. \label{confusion_matrix}}
\end{figure}

We first evaluated the classifier's decision behavior using its pretrained weights under two test configurations: (i) a \emph{Real set} containing real benign ($n=360$) and real melanoma ($n=1061$) images, and (ii) a \emph{Synthetic set} where the benign subset is identical but the melanoma images are replaced by 1000 randomly sampled StyleGAN2-generated images. Figure~\ref{confusion_matrix} summarizes the results.

On the Real set, the classifier achieves near-ceiling performance: all 360 benign images are correctly classified, while 98.8\% (1048/1061) of real melanomas are correctly identified, establishing a strong baseline within our test distribution. On the Synthetic set, melanoma sensitivity decreases to 83.3\% (833/1000), with 16.7\% of synthetic melanomas misclassified as benign. Nevertheless, the majority of generated samples are recognized as melanoma by this strong real-trained model.

This frozen-classifier evaluation directly measures whether synthetic melanoma images fall within the melanoma-relevant decision regions learned from real dermoscopic data, providing task-level evidence that the generated images preserve discriminative disease cues. This complements feature-space metrics (FID, FMD), which assess distributional similarity but not necessarily diagnostic relevance. Importantly, recognizability under a fixed classifier is a necessary---though not sufficient---condition for augmentation utility: if synthetic images are not recognized as melanoma by a strong real-trained model, they are unlikely to improve downstream training when used for data augmentation.

\subsection{Downstream Evaluation II: Augmentation Utility}

Motivated by the recognizability results, we further evaluated whether StyleGAN2-generated melanoma images can improve classifier performance when used as training augmentation. We trained the EfficientNet-B0 classifier at $512 \times 512$ resolution under two controlled training regimes:

\begin{itemize}
    \item Real-only: The model is trained exclusively on real images ($n \approx 20{,}000$), resulting in a highly imbalanced class distribution with a benign-to-melanoma ratio of approximately 98:2.
    \item Real + Synthetic: The training set combines all real images with 6500 StyleGAN2-generated synthetic melanoma images, yielding a more balanced benign-to-melanoma ratio of approximately 65:35.
\end{itemize}

Both training sets were drawn from ISIC 2018 and ISIC 2020, with a random 80/10/10 split for training, validation, and testing. The test set ($n=2000$; 1960 benign, 40 melanoma) consisted exclusively of real images and was held out from all GAN training to ensure unbiased evaluation. Both classifiers were trained for 20 epochs using identical hyperparameters (Adam optimizer, learning rate $10^{-4}$, batch size 32), with the best checkpoint selected based on validation AUC.

Table \ref{tab:classifier_results} reports the results. The model trained with Real + Synthetic data achieved 98.27\% overall accuracy and a melanoma AUC of 0.9445, compared to 85.07\% accuracy and AUC of 0.9252 for the Real-only model. The F1 score for melanoma detection improved from 0.1682 to 0.2586.

\begin{table}[H]
\caption{Classifier performance on the held-out real-image test set. The test set contains 1960 benign and 40 melanoma images, reflecting real-world class imbalance. \label{tab:classifier_results}}

\begin{tabularx}{\textwidth}{lCCC}
\toprule
\textbf{Training Data} & \textbf{Accuracy (\%)} $^\dagger$ & \textbf{Melanoma AUC} & \textbf{Melanoma F1} \\
\midrule
Real-only & 85.07 & 0.9252 & 0.1682 \\
Real + Synthetic & 98.27 & 0.9445 & 0.2586 \\
\bottomrule
\end{tabularx}
\noindent{\footnotesize{$^\dagger$~Accuracy is dominated by the benign majority class; AUC is more informative for imbalanced data.}}
\end{table}

Several considerations apply when interpreting these results. First, overall accuracy is dominated by the benign majority class (98\% of test samples), making it a poor measure of melanoma detection ability; melanoma AUC is more informative as it measures the model's ability to rank melanomas above benign samples across all decision thresholds. Second, the relatively low F1 scores (even after augmentation) reflect the extreme class imbalance in the test set: with only 40 melanoma samples, even a small number of false negatives or false positives substantially impacts precision and recall. Third, the improvement in AUC from 0.9252 to 0.9445 represents a meaningful gain, though formal statistical testing (e.g.,~DeLong's test) would require a larger melanoma test set for adequate power.

Taken together, these results support the claim that StyleGAN2-generated melanoma images are not only recognizable by a strong evaluator but can also provide measurable downstream utility when used to address class imbalance in a controlled training setup.

\section{Dermatologist Evaluation}

To assess the perceptual realism of GAN-generated melanoma images, we constructed a balanced evaluation set of 200 images consisting of 100 real melanomas (randomly sampled from ISIC 2018) and 100 synthetic melanomas (randomly sampled from StyleGAN2 outputs at $\gamma=0.8$). Images were presented in randomized order without any identifying information. We report classification accuracy for (i) a machine baseline and (ii) two board-certified dermatologists who performed the task independently.

\subsection{Machine Baseline: StyleGAN2 Discriminator}
As an initial reference point, we evaluated the trained StyleGAN2 discriminator on the real-versus-synthetic classification task. The discriminator achieved an overall accuracy of 59.5\% (\cref{perf_human_disc}), only modestly above chance (50\%). Notably, the discriminator exhibited an asymmetric error pattern: it achieved 84.0\% accuracy on synthetic images but only 35.0\% on real images, indicating a bias toward classifying images as synthetic. This suggests that the generated images are sufficiently close to the training distribution that even the model's internal real/fake signal provides limited separability. We report the discriminator results \emph{not} as an additional rater, but as a computational benchmark for contextualizing human performance under the same decision setting.


\begin{table}[H]
\caption{Performance comparison of human raters and the StyleGAN2 discriminator on the real-versus-synthetic classification task ($n=200$ images). \label{perf_human_disc}}

\begingroup
\small              
\setlength{\tabcolsep}{4pt} 

\begin{tabularx}{\textwidth}{lCCCC}
\toprule
\textbf{Metric} & \textbf{Dermatologist 1} & \textbf{Dermatologist 2} & \textbf{Human Mean} & \textbf{Discriminator} \\
\midrule
Overall Accuracy & 71.0\% (\emph{p} < 0.001) & 62.0\% (\emph{p} < 0.001) & 66.5\% & 59.5\% \\
Real Accuracy & 51.0\% & 70.0\% & 60.5\% & 35.0\% \\
Synthetic Accuracy & 91.0\% & 54.0\% & 72.5\% & 84.0\% \\
Accepted as Real $^\dagger$ & 9.0\% & 46.0\% & 27.5\% & 16.0\% \\
\bottomrule

\end{tabularx}
\noindent{\footnotesize{$^\dagger$ Percentage of synthetic images classified as real.}}
\endgroup
\end{table}

\subsection{Independent Dermatologist Assessment}

Two board-certified dermatologists independently labeled the 200-image set. Both raters are co-authors of this study and are affiliated with separate major academic medical centers, ensuring independent clinical perspectives. Neither rater had prior exposure to the synthetic images or knowledge of the real/synthetic ratio.

Binomial testing confirmed that both raters performed significantly above the 50\% chance level (Dermatologist~1: 71.0\%, $p < 0.001$; Dermatologist~2: 62.0\%, $p < 0.001$; mean overall accuracy, 66.5\%),
indicating that their selections were deliberate rather than random, despite the difficulty in distinguishing synthetic from real melanoma images.

Notably, the two raters exhibited complementary decision tendencies: Dermatologist 1 achieved high accuracy on synthetic images (91.0\%) but near-chance accuracy on real images (51.0\%), consistent with a conservative strategy that preferentially flags images as synthetic. Dermatologist 2 showed the opposite pattern, with higher accuracy on real images (70.0\%) but lower accuracy on synthetic images (54.0\%), reflecting a more liberal threshold. These complementary labeling patterns indicate that classification difficulty is not confined to a single class and that consistent ``tell-tale'' artifacts are not present across synthetic samples.

\textls[-15]{The percentage of synthetic images accepted as real by each rater was 9.0\% for Dermatologist 1 and 46.0\% for Dermatologist 2, with a mean of 27.5\%. This variability further underscores the absence of consistent visual markers distinguishing synthetic from real~images.}

\subsection{Inter-Rater Reliability}
To quantify agreement beyond chance, we computed Cohen's $\kappa$, defined as~\cite{cohen1960coefficient}:
\[
\kappa = \frac{P_o - P_e}{1 - P_e},
\]
where $P_o$ is the observed agreement and $P_e$ is the expected agreement under chance, given the raters' marginal label distributions. The resulting $\kappa$ values are shown in Table~\ref{cohenk_rater}. Statistical significance of $\kappa$ was assessed using a Z-test, where $Z = \kappa / SE$ and the standard error is given by:
\[
SE = \sqrt{\frac{P_o(1 - P_o)}{n(1 - P_e)^2}}.
\]

\begin{table}[h]
\caption{Inter-rater agreement (Cohen's $\kappa$) for the real-versus-synthetic classification task. \label{cohenk_rater}}
\centering
\begin{tabularx}{\textwidth}{lCCC}
\toprule
\textbf{Comparison} & \textbf{Cohen's $\kappa$} & \textbf{\emph{p}-Value} & \textbf{Agreement} \\
\midrule
Dermatologist 1 vs.\ Dermatologist 2 & 0.173 & 0.009 & Slight \\
Dermatologist 1 vs.\ Discriminator & 0.042 & 0.482 & Slight \\
Dermatologist 2 vs.\ Discriminator & 0.082 & 0.187 & Slight \\
\bottomrule
\end{tabularx}
\end{table}

Inter-rater agreement between the two dermatologists was low but statistically significant ($\kappa = 0.173$, $p = 0.009$), indicating substantial variability in labeling criteria even among experts \cite{landis1977measurement}. Agreement between each dermatologist and the discriminator was negligible and not statistically significant ($\kappa \leq 0.082$, $p > 0.05$), consistent with humans and the discriminator relying on different visual cues.

\subsection{Summary}
Both machine and human evaluations converge on the same conclusion: distinguishing StyleGAN2-generated melanoma images from real melanomas is difficult under visual inspection. The modest above-chance accuracy (66.5\% human mean), low inter-rater agreement ($\kappa = 0.173$), and complementary response patterns across raters collectively support the perceptual realism of the generated samples.

\section{Limitations and Future Work}

Certain tradeoffs of the analyzed generators should be noted. DCGAN exhibits limited capacity for high-resolution melanoma synthesis, often failing to preserve fine-grained dermoscopic details. StyleGAN3-T shows limited mode coverage, producing individually realistic samples while failing to capture the full diversity of melanoma appearances. StyleGAN3-R improves distributional fidelity but introduces mesh- or grid-like artifacts that are undesirable in medical images. StyleGAN2 achieves strong overall performance but remains sensitive to regularization settings.

Future work will extend this study in several directions. First, evaluating cross-dataset generalization to other imaging modalities (e.g., smartphone-captured images) and external datasets will assess the robustness of synthetic augmentation strategies. Second, given recent advances in diffusion-based generative models, a comparative evaluation of latent diffusion models against the GAN architectures benchmarked here will determine whether these newer approaches offer advantages for preserving fine-grained dermoscopic features. Third, extending to conditional generation would address specific gaps in training data, such as skin type, melanoma subtype, and anatomical location. 

Finally, structured expert assessment using the 7-point dermoscopic checklist will validate clinical feature preservation and identify artifacts not captured by automated metrics. Additionally, integrating synthetic images with automated 7-point checklist detection systems \cite{wadhawan2011implementation, zouridakis2020methods, lancaster2023asymmetry} could determine whether GAN-generated melanomas preserve the specific dermoscopic features (e.g., atypical pigment network, asymmetry, vascular patterns) required for algorithmic classification.

\section{Conclusions}

This study presents the first systematic benchmark of four GAN architectures for high-resolution ($512 \times 512$) melanoma image synthesis, addressing a critical bottleneck in dermatological AI: the scarcity of annotated melanoma images and severe class imbalance in training datasets. Using consistent protocols and multi-faceted evaluation on two expert-annotated benchmarks (ISIC 2018 \cite{Codella2019ISIC2018Challenge} and ISIC 2020 \cite{Rotemberg2021}), we demonstrate that StyleGAN2 achieves the optimal balance of distributional fidelity, perceptual realism, and artifact avoidance, attaining FID scores of 24.8 and 7.96 on the respective datasets.

Three lines of evidence support the diagnostic relevance of StyleGAN2-generated melanomas: (1) a frozen EfficientNet-based classifier recognized 83\% of synthetic samples as melanoma, confirming preservation of disease-discriminative features; (2) board-certified dermatologists from independent institutions distinguished synthetic from real images at only 66.5\% accuracy, demonstrating the absence of consistent visual artifacts; and (3)~augmenting a class-imbalanced training set with synthetic melanomas improved detection AUC from 0.925 to 0.945, providing direct evidence of downstream clinical utility.

These results demonstrate that high-quality synthetic melanoma images can serve as a practical tool for mitigating class imbalance in melanoma detection pipelines. As melanoma remains the deadliest form of skin cancer, with outcomes highly dependent on early detection, methods that improve automated screening systems have significant potential clinical impact. Moreover, the proposed framework may extend to other data-scarce dermatological conditions, such as Buruli ulcer disease, where the limited availability of annotated images similarly hinders the development of reliable automated screening tools~\cite{queen2023towards}. This work establishes a foundation for integrating synthetic data into dermatological AI development---not as a substitute for real patient data, but as a complementary resource for improving model robustness and generalization.

\vspace{6pt}


\authorcontributions{Conceptualization, G.Z.; methodology, P.Y.L., R.H. and S.H.; software, P.Y.L., Y.S., N.M. and S.Y.; algorithm validation, Y.S.; clinical validation, C.E.W., A.C.; formal analysis, N.M. and R.H.; investigation, C.M.Q. and G.Z.; resources, Y.S. and N.M.; data curation, P.Y.L. and N.M.; writing – original draft, P.Y.L., G.Z.; writing – review and editing, Y.S., N.M., R.H., and G.Z.; supervision, R.H. and G.Z.; project administration, G.Z.; funding acquisition, C.M.Q. and G.Z. All authors have read and agreed to the published version of the manuscript.
}

\funding{This work was partially supported by the Texas Tech University Innovation Hub, Prototype Fund, 2022.}

\dataavailability{ The datasets used in this study are publicly available from the ISIC archive \cite{Rotemberg2021, Codella2019ISIC2018Challenge}.
}

\conflictsofinterest{The authors declare no conflicts of interest. 
} 

\begin{adjustwidth}{-\extralength}{0cm}

\reftitle{References}

\PublishersNote{}

\end{adjustwidth}

\end{document}